# Incremental Centering and Center Ambiguity

**Udo Hahn & Michael Strube**

Computational Linguistics Research Group
Freiburg University
Europaplatz 1
D-79085 Freiburg, Germany
{hahn,strube}@coling.uni-freiburg.de

**Abstract**

In this paper, we present a model of anaphor resolution within the framework of the centering model. The consideration of an incremental processing mode introduces the need to manage structural ambiguity at the center level. Hence, the centering framework is further refined to account for local and global parsing ambiguities which propagate up to the level of center representations, yielding moderately adapted data structures for the centering algorithm.

## Introduction

Psycholinguistic studies have revealed ample evidence for the incrementality of human language comprehension, not only at the phrasal and clausal level but also at the discourse level of anaphora resolution (Just & Carpenter, 1987; Sanford & Garrod, 1989). Correspondingly, incremental processing has also become a major challenge for cognitively plausible, computational models of natural language understanding (Jurafsky, 1992; Sturt, 1995), and text understanding (Granger et al., 1986) in particular. Introducing incrementality into the centering model (Grosz et al., 1995), the methodological framework for our approach to the resolution of (pro)nominal anaphora, however, is not at all straightforward. In particular, incremental processing introduces (local) ambiguities at significant rates, which cannot be properly accounted for at the center level in the original model. Though centering strives for the elimination of *referential* ambiguities, the implications of *structural* ambiguities have been completely ignored so far.

We have gathered some data, summarized in Table 1, to give an empirical assessment of the relevance of the issue under investigation. Altogether 47 texts (product reviews from the information technology domain) were analyzed which consist of 32291 words, with 230 occurrences of (un)ambiguous pronouns.

| ambiguous | 174 | (76 %) | | |
|---|---|---|---|---|
| locally | | | 145 | (63 %) |
| globally | | | 29 | (13 %) |
| unambiguous | 56 | (24 %) | | |

Table 1: Ambiguity Distribution Patterns of Pronouns

Given our text corpus, 87% of the sentences could have been processed by the original, non-incremental centering algorithm (only global ambiguities could not). This rate drops dramatically to only 24% when we assume an incremental operation mode. These data reflect the impact of the local ambiguities which are resolved at the sentence level as the parse proceeds and thus are not an issue for the original (non-incremental) centering algorithm. The latter percentage rate, however, gives a realistic picture of the relevance of the problem under scrutiny when one opts for a cognitive adequate, incremental model of text understanding.

## Brief Survey of the Centering Model

The framework of our model of anaphora resolution is provided by the well-known *centering* mechanism (Grosz et al., 1995), for which psycholinguistic evidences are provided by Gordon et al. (1993) and Brennan (1995), lacking, however, the consideration of incrementality of language processing. The theory of centering is intended to model the local coherence of discourse, i.e., coherence among the utterances $U_i$ in a particular discourse segment (say, a paragraph of a text). Local coherence is opposed to global coherence, i.e., coherence with other segments in the discourse. Discourse entities serving to link one utterance to other utterances in a particular discourse segment are organized in terms of centers. Each utterance $U_i$ in a discourse segment is assigned a set of *forward-looking centers*, $C_f(U_i)$, and a unique *backward-looking center*, $C_b(U_i)$. The forward-looking centers of $U_i$ depend only on the expressions that constitute that utterance, previous utterances provide no constraints on $C_f(U_i)$. The elements of $C_f(U_i)$ are partially ordered to reflect relative prominence in $U_i$. The most highly ranked element of $C_f(U_i)$ that is *realized* in $U_{i+1}$ (i.e., is associated with an expression that has a valid interpretation in the underlying semantic/conceptual representation language) is the $C_b(U_{i+1})$. The ranking imposed on the elements of the $C_f$ reflects the assumption that the most highly ranked element of $C_f(U_i)$ is the most preferred antecedent of an anaphoric expression in $U_{i+1}$, while the remaining elements are (partially) ordered according to decreasing preference for establishing referential links.

The theory of centering, in addition, defines several transition relations across pairs of adjacent utterances (e.g., continuation, retainment, smooth and rough shift), which differ from each other according to the degree by which successive backward-looking centers are confirmed or rejected, and, if they are confirmed, whether they correspond to the most highly ranked element of the current forward-looking centers or not. The theory claims, above all, that to the extent a discourse adheres to all these centering constraints (e.g., realization constraints on pronouns, preferences among types of center transitions), its local coherence will increase and the inference load placed upon the hearer will decrease. Therefore, the tremendous importance of fleshing out the relevant and most restrictive, though still general centering constraints.

## Incremental Centering and Ambiguity

In this section, we argue for an extension of the centering model that accounts for ambiguities generated by the incremental operation of the parsing component of a text understanding system. Though we also provide for mechanisms that deal with global structural ambiguities, we here concentrate on local structural ambiguities in the phrase which contains an anaphor. These can be directly attributed to the *incremental* processing mode, where each lexical element is integrated in syntactic structures and semantically interpreted as early as possible. As the anaphora resolution is also executed incrementally, *local* syntactic ambiguities (which cause different referential entities to emerge at the semantic/conceptual level of interpretation) must be accessible through the data structures of the centering algorithm in order to maintain local, alternative center readings.

Consider the text fragment ((1) – (3)) taken from the corpus of product reviews:

(1) In der Leistung konnte *die LPS 105* ebenfalls weitgehend überzeugen.
    (As far as performance is concerned, *the LPS 105* also produced rather compelling results.)

(2) Bei der mittleren Zugriffszeit (16,5 ms) erreicht *diese Festplatte die Seagate ST-3144*, womit *sie* in dieser Disziplin den zweiten Platz erzielt.
    (Regarding the mean access time (16,5 ms) *this hard disk* compares to the *Seagate ST-3144*, by which *it* scores second-best in this category.)

(3) Auch beim Datendurchsatz erweist *sie* sich als hochkarätiges Produkt.
    (Also, considering data throughput *it* turns out to be a high-caliber product.)

Sentence (1) has a unique structural analysis, the *forward-looking centers* ($C_f$) consist of two semantic/conceptual elements, the LPS-105 hard disk and PERFORMANCE (cf. Table 2). In sentence (2), a nominal anaphor occurs, *"diese Festplatte" (this hard disk)*, which is resolved to LPS-105 from the previous sentence. Unfortunately, the noun phrase *"diese Festplatte"* is nominative as well as accusative and may be alternatively attached to the verb *"erreicht" (compares to)* both in its subject and object role (we assume a dependency grammar framework as briefly described in the following section). In this state, one cannot determine which of the grammatical functions is the correct one, thus a structural ambiguity has been identified. Since the second NP in this sentence *("die Seagate ST-3144")* is ambiguous with respect to both of these cases, too, the parser produces two structurally and conceptually ambiguous readings (with inverted subject/object instantiations; given appropriate stress marking both readings are equally plausible). As a consequence, two different $C_f$s[1] have to be created (cf. Table 2), which indicate two different center transitions, *viz.* continuation vs. retention, eligible at the end of the analysis of the second sentence. This choice option becomes crucial for the resolution of the pronoun *"sie" (it)* in sentence (3), as it depends on the appropriate selection of one of the two different $C_f$s. In the case of the CONTINUE transition (the $C_b$ of the previous utterance is also the highest ranked element of the $C_f$s of the current utterance) LPS-105 is preferred as the antecendent, while in the case of the RETAIN transition (the $C_b$ of the previous utterance is not the highest ranked element of the $C_f$s of the current utterance) it is ST-3144. Depending on how the text actually proceeds either one is equally possible. So, for the actual anaphora resolution the transition type preferences (Rule 2 in Grosz et al. (1995)) are of no help at all to decide among any of these variants. We, therefore, conclude that additional representation devices have to be supplied to keep track of these structurally induced ambiguities at the center level.

| (1) | **Cb:** | LPS-105: LPS 105 | |
| | **Cf:** | [LPS-105: LPS 105, PERFORMANCE: Leistung] | CONTINUE |
| (2) | **Cb1:** | LPS-105: Festplatte | |
| | **Cf1:** | [LPS-105: Festplatte, ST-3144: Seagate ST-3144, ACCESS-TIME: Zugriffszeit, RANK: Platz, CATEGORY: Disziplin] | CONTINUE |
| | **Cb2:** | LPS-105: Festplatte | |
| | **Cf2:** | [ST-3144: Seagate ST-3144, LPS-105: Festplatte, ACCESS-TIME: Zugriffszeit, RANK: Platz, CATEGORY: Disziplin] | RETAIN |

Table 2: Centering Data for Sentences (1) and (2)

## Grammar Constraints on Anaphora

We now consider several constraints on anaphora which apply both to the sentence-level and text-level of anaphora analysis. These descriptions will later serve as a framework for considering local ambiguity within the centering approach. We here adapt the common binding criteria to the methodological requirements of a fully lexicalized dependency grammar (DG), introducing the central notion of *d-binding*[2] (cf. Strube

---

[1] To simplify the presentation, we will assume the canonical ordering on $C_f$ based on grammatical roles, *viz.* SUBJECT > OBJECT(S) > OTHERS (Grosz et al., 1995, p.214). We have clear evidences, whatsoever, that this is inappropriate, for German and related free word order languages at least, and argue for ordering criteria based on the functional information structure of utterances in terms of topic/comment or theme/rheme patterns in a companion paper (Strube & Hahn, 1996).

[2] The definition of *d-binding* (cf. Table 3) corresponds to the *governing category* in GB terminology, which relies upon the notion of

& Hahn (1995) for an elaborated discussion of *d-binding* criteria). These constraints hold for intra- as well as intersentential anaphora, thus seamlessly incorporating the discourse level of grammatical description (for a comprehensive survey of the grammar formalism, cf. Hahn et al. (1994)).[3]

The possible antecedents that can be reached via anaphoric relations, irrespective of whether they occur within the current sentence or beyond, are described by *isPotentialAnaphoricAntecedentOf* (cf. Table 4), which incorporates *d-binds* (cf. Table 3).

```
x d-binds y :⇔
(x head⁺ y)
∧ ¬∃ z: ((x head⁺ z) ∧ (z head⁺ y)
    ∧ (z isa_C* finiteVerb
    ∨ ∃u: (z head u
        ∧ ((z spec u ∧ u isa_C* DetPossessive)
        ∨ (z saxGen u ∧ u isa_C* Noun)
        ∨ (z ppAtt u ∧ u isa_C* Noun)
        ∨ (z genAtt u ∧ u isa_C* Noun)))))
```

Table 3: D-binding Constraint

```
x isPotentialAnaphoricAntecedentOf y :⇔
¬∃ z: (z d-binds x ∧ z d-binds y)
∧ ((∃ u: u d-binds y ∧ u head⁺ x) → x left⁺ y)
```

Table 4: Constraint on Potential Anaphoric Antecedents

*PronAnaphorTest* from Table 5 contains the major grammatical agreement constraint (covering gender, number and person) for some anaphoric pronoun and its nominal antecedent, while *NomAnaphorTest* from Table 6 captures the major conceptual subsumption constraint for the nominal antecedent and a corresponding anaphoric definite NP.

---

*c-command*, while the DG constraint on anaphora (cf. Table 4) relates to the major binding principles of GB (Chomsky, 1981). An approach to the incremental computation of intrasentential coreferences based on Chomsky's binding theory is given by Merlo (1993).

[3] For the definitions of the grammatical predicates below, the following conventions hold: $isa_C$ denotes the subclass relation among lexical classes (parts of speech), ⊔ the unification operation, ⊥ the inconsistent element. Let $u$ be a complex feature term and $l$ a feature name, then the extraction $u \backslash l$ yields the value of $l$ in $u$. If $l$ is defined, $u \backslash l$ gives ⊥ in all other cases. Semantic and conceptual knowledge is represented via a KL-ONE-style classification-based knowledge representation language (MacGregor & Bates, 1987), with $isa_F$ denoting the subclass relation among concepts. Furthermore, *object.attribute* denotes the value of the property *attribute* at *object* and the symbol self refers to the current lexical item. The grammar specification language, in addition, incorporates topological primitives for relations within dependency trees, such as "*x* occurs left of *y*" and "*x* is head of *y*"; rel⁺ and rel* denote the transitive and transitive/reflexive closure of a relation rel, respectively. The following dependency relations will be used: specifier-of (spec), Saxon genitive (saxGen), prepositional and genitival attribute (ppAtt, genAtt).

```
PronAnaphorTest (pro, ante):⇔
  ante isa_C* Nominal ∧
  ((pro.features\self\agr\gen)
    ⊔ (ante.features\self\agr\gen) ≠ ⊥) ∧
  ((pro.features \self\agr\num)
    ⊔ (ante.features\self\agr\num) ≠ ⊥) ∧
  ((pro.features\self\agr\pers)
    ⊔ (ante.features\self\agr\pers) ≠ ⊥)
```

Table 5: Constraint on Pronominal Anaphora

```
NomAnaphorTest (defNP, ante):⇔
  ante isa_C* Nominal ∧
  ((defNP.features \self\agr\num)
    ⊔ (ante.features\self\agr\num) ≠ ⊥) ∧
  ante.concept isa_F* defNP.concept
```

Table 6: Constraint on Nominal Anaphora

## Resolution of Anaphora

The actor computation model (Agha & Hewitt, 1987) provides the background for the procedural interpretation of lexicalized grammar specifications, as those given in the previous section, in terms of so-called word actors. Word actors communicate via asynchronous message passing; an actor can only send messages to other actors it knows about, its so-called acquaintances. The arrival of a message at an actor triggers the execution of a method that is composed of grammatical predicates (for a survey, cf. Neuhaus & Hahn (1996)).

The basic data structures for anaphora resolution are organized as acquaintances of specific actors. Besides word actors for the lexical level of analysis, phrases are encapsulated in *PhraseActors*, and one or more *PhraseActors* which cover the same sequence of words but assign different syntactic interpretations (local ambiguities) to it are encapsulated in *ContainerActors*. For every sentence, its associated unique *ParserActor* is acquainted with a *CenteringActor* which, for reasons of ambiguity handling, is acquainted with one or more *CenterActors*. Each of these *CenterActors* has a preferentially ordered list of forward-looking centers ($C_f$) and a single backward-looking center ($C_b$). The usual criteria for centering apply at this representation level (Grosz et al., 1995). We extend this basic model, however, in that we provide several instances of *CenteringActors* to account for local ambiguities within an utterance, while different *CenterActors* represent global ambiguities of single utterances. Hence, unambiguous centering is a special case, where a single *CenteringActor* is only acquainted with a single *CenterActor*.

Anaphora analysis encompasses the procedural interpretation of the declarative constraints given in the previous section. For *pronominal* anaphors, the *SearchPronAntecedent* message is triggered by the successful syntactic test that the pronoun may be modifier of its head. For *nominal* anaphors, the *SearchNomAntecedent* message is triggered by the attachment of a definite determiner as a modifier to its head noun and a similar syntactic head-modifier compatibility test[4]. We now describe the main threads of the algorithm for local ambiguity management during anaphora resolution (cf. Fig. 1).

Consider, e.g., sentence (2) of the already introduced text fragment. An attachment of the definite NP *"diese Festplatte" (this hard disk)* at its prospective head *"erreicht" (compares to)* is tried. Since the NP is ambiguous with respect to case, a local ambiguity accounting for the subject and object reading is created[5].

**1, 1a:** Two different *SearchNomAntecedent* messages with the argument *theAttachment* (the dependency relation between the NP and *"erreicht"*, i.e., either `subject` or `object`) are sent simultaneously from the *Anaphor* to the *PhraseActor*.

**2–4, 2a–4a:** Both messages are forwarded from the *PhraseActor* to the *ContainerActor*, the *ParserActor*, and the *CenteringActor* of the preceding sentence.

**5, 5a:** It is crucial that for every *SearchAntecedent* message which reaches the *CenteringActor*, that actor is copied, leaving the master copy unchanged. This guarantees that each locally ambiguous phrase which contains an anaphor manipulates its own centering data structures.

**6, 6a:** The messages are distributed to all *CenterActors* where the argument *theAttachment* is copied in order to provide for consistent data in a distributed, concurrent environment (the copy action is of relevance only in those cases where global and local ambiguities are interleaved).

**7, 7a:** *NomAnaphorTest* succeeds in both cases (the most preferred element of the $C_f$ of sentence (1), viz. LPS-105 fulfills the required conceptual subsumption condition relative to *"Festplatte" (hard disk)*).

**8, 8a:** Hence two *AntecedentFound* messages are sent to each corresponding anaphor.

**9, 9a:** The semantic predicate *permit*[6] succeeds with respect to the word actors *LPS 105* (*Festplatte*) and *erreicht* for both argument positions, thus the dependency relations *theAttachment*, viz. `subject` and `object`, are confirmed.

**10, 10a:** The resolved anaphors send an *AnaphorSucceed* message to the corresponding *CenterActors*.

**11, 11a:** At these *CenterActors* the determined antecedents (LPS-105) are removed from the corresponding $C_f$ lists. The removal of the antecedent from the $C_f$ list prevents it from being (incorrectly) reused as a possible antecedent for yet another anaphor within the *same* sentence.

As a special case of local ambiguities, consider the second anaphor in sentence (2), the pronoun *"sie" (it)*. Caused by the attachment of *"sie"* to its prospective head *"erzielt" (scores)* in the subordinate clause of sentence (2), and due to the case ambiguity of *"sie"*, viz. nominative and accusative, the corresponding *PhraseActors* are duplicated (as the matrix clause is ambiguous, too, four interpretations must be considered; in the corresponding Fig. 2 only two readings are shown). Four *SearchAntecedent* messages are triggered. Steps **1–8** are performed as described above. As the corresponding $C_f$s of (1) contain only PERFORMANCE (LPS-105 has already been consumed as a result of previously resolving *"diese Festplatte"*), the predicate *permit* fails with respect to PERFORMANCE and *"erzielt"* (Step **9**).

**12, 12a:** The anaphor sends an *AnaphorReject* message to the *CenterActor*.

**13, 13a:** The $C_f$ list is exhausted.

**14, 14a:** Hence, the mechanism for intrasentential anaphora resolution is triggered. The search for an antecedent is performed <u>within</u> the *PhraseActor* which contains both the anaphor and the antecedent (cf. Fig. 2; it differs from Fig. 1 mainly with respect to description of the the dependency structures within each *PhraseActor*, which are depicted in greater detail).

**15, 15a:** Each *SearchAntecedent* message is forwarded from its initiator *"sie"* to its prospective head *Head1 "erzielt" (scores)* which d-binds the initiator.

**16, 16a:** Next the message is forwarded to *Head2 "erreicht" (compares to)*.

**17, 17a:** Then the message is forwarded to possible *Antecedent*s which are modifiers of *Head2*, where *"diese Festplatte" (this hard disk)* and *"ST-3144"* are reached, respectively.

**18, 18a:** *PronAnaphorTest* and the semantic predicate *permit* succeed (*"diese Festplatte"*, which is resolved to LPS-105, and *"erzielt"* as well as ST-3144 and *"erzielt"* are both successfully tested by *permit*).

**19, 19a:** An *AntecedentFound* message is sent to the anaphor. The dependency relation *theAttachment*, viz. `subject`, is confirmed in both cases between *"erzielt" (scores)* and LPS-105 as well as ST-3144, respectively. Similarly, the `object` dependency relation ist established, until the accusative phrase *"den zweiten Platz" (second-best in this category)* invalidates this local ambiguity.

Upon completion of the analysis of sentence (2) two *CenterActors* continue to exist with corresponding $C_b/C_f$ data

---

[4]We strictly separate the search for the proper antecedent and the evaluation of its conceptual compatibility as the modifier of a head.

[5]The bold numbers in the text and the edge numbers in Fig. 1 and 2 refer to the same computation steps. The index *a* indicates parallel distribution of messages. The directed edges in both figures illustrate the basic flow of control caused by the message passing.

[6]*permit* accounts for type and further conceptual admissibility constraints (number restrictions, etc.).

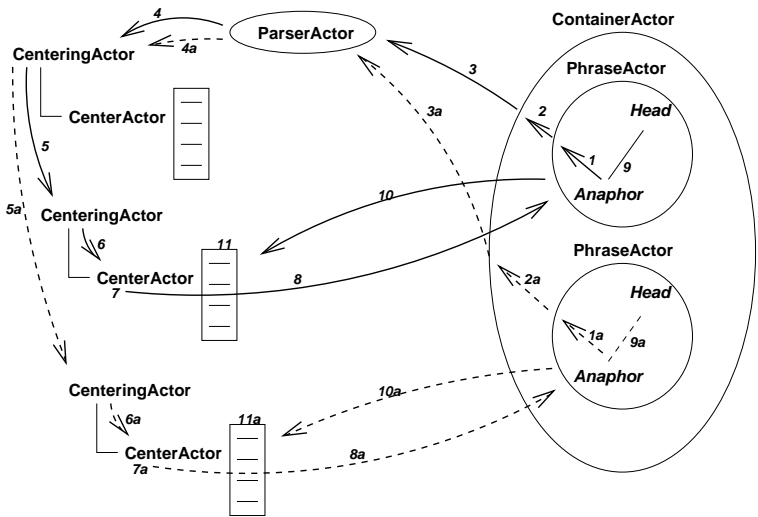

Figure 1: Protocol for the Local Ambiguity Case at the Text Level

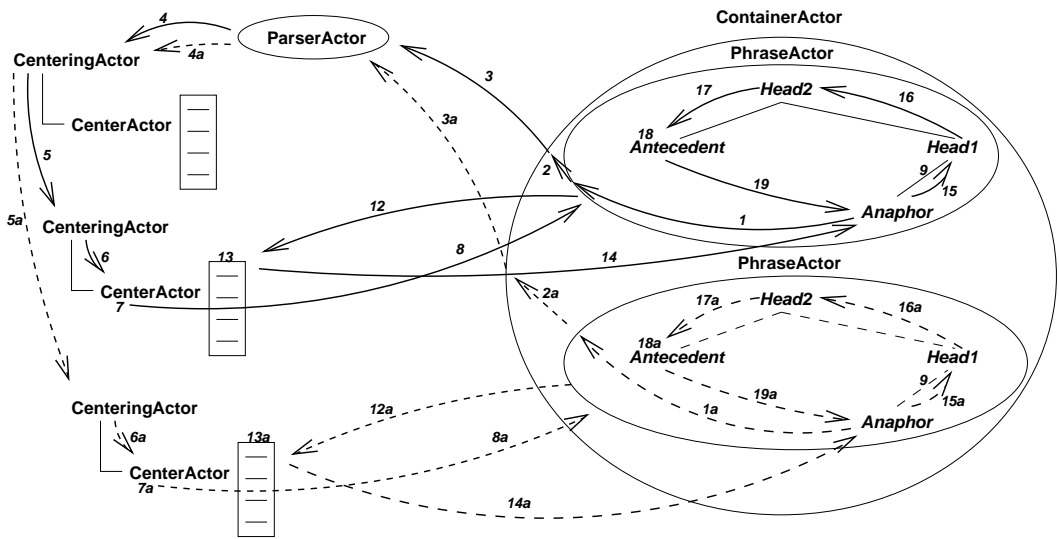

Figure 2: Protocol for the Local Ambiguity Case at the Phrase Level

(cf. Table 2). Hence, the local ambiguity with respect to *"sie"* no longer persists and has been reduced to a global one.

Summarizing, we propose a two-level representation of structural ambiguities for the centering model, one at which local and global structural ambiguities are made explicit. Global ambiguities are represented as sets of forward-looking centers (a so-called *centering set*, in the underlying implementation realized as *CenterActor*), while local ambiguities are represented as a set of such centering sets (in the underlying implementation realized as *CenteringActor*). The creation and management of these sets is under control of the parser, while the management of entities within these centers remains in the realm of the centering theory. The proposal we make does not depend on any choice of the underlying grammar or semantic theory (although binding criteria should be expressable).

## Related Work

The centering model, from its inception (Grosz et al., 1983) to its most recent formulation (Grosz et al., 1995), has been considered a methodological framework for anaphora resolution. With the exception of Brennan et al. (1987), whose implementation was interfaced with a concrete HPSG system, the centering approach seems to have been developed as a stand-alone theory vehicle, with almost no attention given to its integration into a larger NLU system framework. This might explain why the issue of *structural* ambiguity handling has been largely ignored in the centering framework. The problem of *referential* ambiguity to which our proposal is equally applicable has recently been discussed by Walker et al. (1994). However, their problem concerns the choice options arising for the assignment of alternative discourse entities from the forward-looking center list to zero-anaphorized arguments of verbs in Japanese, given that the number of candidate antecedents exceeds that of the zero anaphors. Their use of functional notions, viz. topicality and empathy, naturally extend the role preferences of the underlying centering model, but unlike our case, <u>no</u> structural ambiguities are involved. In contrast, our approach applies to the representation of both types of ambiguities.

We motivated the treatment of (structural) ambiguities within the centering framework as a consequence of assuming an *incremental* mode of anaphor resolution, a topic that has not been raised in the centering literature so far. This is surprising insofar as even psycholinguistic studies on centering (Gordon et al., 1993; Brennan, 1995) do not touch upon this issue, though the immediacy of anaphor resolution is a common theme in cognitive text processing studies (Just & Carpenter, 1987; Sanford & Garrod, 1989).

Our proposal, based on a dependency-style grammar model (Hahn et al., 1994), claims to integrate both the sentence-level as well as text-level of anaphora analysis. Furthermore, it is also fully integrated with terminological reasoning facilities as needed for in-depth text understanding, and is based on an incremental, single-pass procedure. Thus, it is superior to the work on binding theory as developed within the GB framework (Chomsky, 1981) that is restricted to the sentence-level of analysis; just recently, however, Merlo (1993) has proposed an incremental procedure for computing intrasentential coreferences based on binding theory constraints. Also Haddock (1987) considers an incremental mode of anaphora resolution which boils down to a variable binding, i.e., a constraint satisfaction problem in the context of a Combinatory Categorial Grammar. Any of these approaches neglects the important aspect of a preference scaling for properly selecting among several candidate discourse units as antecedents. This drawback in the same way applies to the framework of DRT (Kamp & Reyle, 1993), which is also non-incremental.

## Conclusions

Our approach to anaphora resolution extends the original centering model by embedding the centering approach into an *incremental*, single-pass processing model, by providing data structures for the centering algorithm which allow for the treatment of local and global (parsing) *ambiguities*, and by homogeneously integrating the resolution of *sentence-level (intrasentential)* as well as *text-level (intersentential)* anaphora based on the strict requirements set up by the *binding criteria* (adapted to a dependency grammar framework).

The anaphora resolution module has been realized as part of a dependency parser for the German language. The parser has been implemented in Actalk (Briot, 1989), an actor language dialect of Smalltalk. The current lexicon contains nearly 3.000 lexical entries and corresponding concept descriptions from two domains (information technology and medicine) available from the LOOM knowledge representation system (MacGregor & Bates, 1987).

**Acknowledgments.** We would like to thank our colleagues in the $\mathcal{CLIF}$ group who read earlier versions of this paper. In particular, improvements are due to discussions we had with Peter Neuhaus. This work has been funded by *LGFG Baden-Württemberg* (1.1.4-7631.0; M. Strube) and a grant from *DFG* (Ha 2907/1-3; U. Hahn).

## References

Agha, G. & C. Hewitt (1987). Concurrent programming using actors. In A. Yonezawa & M. Tokoro (Eds.), *Object-Oriented Concurrent Programming*, pp. 37–53. Cambridge, Mass.: MIT Press.

Brennan, S. E. (1995). Centering attention in discourse. *Language and Cognitive Processes*, 10(2):137–167.

Brennan, S. E., M. W. Friedman & C. J. Pollard (1987). A centering approach to pronouns. In *Proc. of ACL-87*, pp. 155–162.

Briot, J.-P. (1989). Actalk: A testbed for classifying and designing actor languages in the Smalltalk-80 environment. In *Proc. of ECOOP-89*, pp. 109–129.

Chomsky, N. (1981). *Lectures on Government and Binding*. Dordrecht: Foris.

Gordon, P. C., B. J. Grosz & L. A. Gilliom (1993). Pronouns, names, and the centering of attention in discourse. *Cognitive Science*, 17:311–347.

Granger, R. H., K. P. Eiselt & J. K. Holbrook (1986). Parsing with parallelism: A spreading-activation model of inference processing during text understanding. In J. Kolodner & C. Riesbeck (Eds.), *Experience, Memory, and Reasoning*, pp. 227–246. Hillsdale, N.J.: L. Erlbaum.

Grosz, B. J., A. K. Joshi & S. Weinstein (1983). Providing a unified account of definite noun phrases in discourse. In *Proc. of ACL-83*, pp. 44–50.

Grosz, B. J., A. K. Joshi & S. Weinstein (1995). Centering: A framework for modeling the local coherence of discourse. *Computational Linguistics*, 21(2):203–225.

Haddock, N. (1987). Incremental interpretation and Combinatory Categorial Grammar. In *Proc. of IJCAI-87*, Vol. 2, pp. 661–663.

Hahn, U., S. Schacht & N. Bröker (1994). Concurrent, object-oriented dependency parsing: The *ParseTalk* model. *International Journal of Human-Computer Studies*, 41(1/2):179–222.

Jurafsky, D. (1992). An on-line computational model of human sentence interpretation. In *Proc. of AAAI-92*, pp. 302–308.

Just, M. & P. Carpenter (1987). *The Psychology of Reading and Language Comprehension*. Boston, Mass.: Allyn & Bacon.

Kamp, H. & U. Reyle (1993). *From Discourse to Logic*. Dordrecht: Kluwer.

MacGregor, R. & R. Bates (1987). *The LOOM Knowledge Representation Language*. (ISI/RS-87-18) USC/ISI.

Merlo, P. (1993). For an incremental computation of intrasentential coreference. In *Proc. of IJCAI-93*, Vol. 1, pp. 1216–1221.

Neuhaus, P. & U. Hahn (1996). Restricted parallelism in object-oriented lexical parsing. In *Proc. of COLING-96*.

Sanford, A. & S. Garrod (1989). What, when, and how?: Questions of immediacy in anaphoric reference resolution. *Language and Cognitive Processes*, 4(3/4):235–262.

Strube, M. & U. Hahn (1995). *ParseTalk* about sentence- and text-level anaphora. In *Proc. of EACL-95*, pp. 237–244.

Strube, M. & U. Hahn (1996). Functional centering. In *Proc. of ACL-96*.

Sturt, P. (1995). Incorporating "unconscious reanalysis" into an incremental, monotonic parser. In *Proc. of EACL-95*, pp. 291–296.

Walker, M. A., M. Iida & S. Cote (1994). Japanese discourse and the process of centering. *Computational Linguistics*, 20(2):193–233.